\documentclass[conference]{IEEEtran}
\IEEEoverridecommandlockouts
\IEEEpubid{\makebox[\columnwidth]{978-1-7281-9023-5/21/\$31.00~\copyright{} 2021 IEEE \hfill}
\hspace{\columnsep}\makebox[\columnwidth]{ }}

\newcommand{\ve}[1]{\mbox{\boldmath $ #1 $}}
\newcommand{\dve}[1]{\dot{\mbox{\boldmath $ #1 $}}}
\newcommand{\ddve}[1]{\ddot{\mbox{\boldmath $ #1 $}}}

\usepackage[pdftex]{graphicx}
\usepackage{cite}
\usepackage{amsmath,amssymb,amsfonts}
\usepackage{algorithmic}
\usepackage{textcomp}
\begin{document}

\title{Bilateral Control-Based Imitation Learning \\
for Velocity-Controlled Robot
\thanks{
This work was supported by JST PRESTO Grant Number JPMJPR1755, Japan. And, this research is also supported by Adaptable and Seamless Technology transfer Program through Target-driven R\&D (A-STEP) from Japan Science and Technology Agency (JST) Grant Number JPMJTR20RG.
}
}

\author{\IEEEauthorblockN{1\textsuperscript{st} Sho Sakaino}
\IEEEauthorblockA{\textit{ Department of Intelligent Interaction Technologies} \\
\textit{ University of Tsukuba and JST, PRESTO}\\
Tsukuba, Japan \\
sakaino@iit.tsukuba.ac.jp}
}

\maketitle
\begin{abstract}
Machine learning is now playing important role in robotic object manipulation.
In addition, force control is necessary for manipulating various objects to achieve robustness against perturbations of configurations and stiffness.
The author's group revealed that fast and dynamic object manipulation with force control can be obtained by bilateral control-based imitation learning.
However, the method is applicable only in robots that can control torque, while it is not applicable in robots that can only follow position or velocity commands like many commercially available robots.
Then, in this research, a way to implement bilateral control-based imitation learning to velocity-controlled robots is proposed.
The validity of the proposed method is experimentally verified by a mopping task.
\end{abstract}


\section{Introduction}
Machine learning is a promising technique to obtain complicated tasks that are difficult to design by humans.
For example, a study using reinforcement learning by Levine \cite{levine17:_learn} \textit{et al.} is one of the leading methods for manipulating objects using machine learning in recent years, and many reinforcement learning-based methods have been reported since then \cite{tsurumine19:_deep}.
However, the learning efficiency of reinforcement learning is very low because the majority of its trials fail, and it is far from the practical level. 
Imitation learning, which can solve these problems, has been a hot topic in recent years \cite{yang16:_repeat_foldin_task_human_robot}\cite{Calinon2007}\cite{kyrarini19:_robot_learn_indus_assem_task_human_demon}\cite{kormushev11:_imitat_learn_posit_force_skill}\cite{zhang18:_deep_imitat_learn_compl_manip}. 
In imitation learning, machine learning such as neural networks are used to infer appropriate actions for a task from multiple actions demonstrated by a human. 
Therefore, it is expected to be practical because of its high sample efficiency, which is basically due to the fact that it learns successful actions in the task in supervised learning frameworks.
Research in imitation learning is being conducted vigorously, including methods that use image information \cite{jiny20:_geomet_persp_visual_imitat_learn}, methods that use force response values to improve adaptability to irregularly shaped objects \cite{lee15:_learn_force_based_manip_defor}\cite{rozo13}\cite{ochi18:_deep_learn_scoop_motion_using_bilat_teleop}\cite{rozo19:_inter_trajec_adapt_force_baysian_optim}\cite{osa18:_onlin_trajec_plann_force_contr}\cite{rozo15:_learn}, and methods that combine reinforcement learning \cite{gupta2019relay}.

However, most of the methods that use machine learning such as reinforcement learning and imitation learning can only perform static actions, and can perform tasks at 1/3 to 1/10 of the speed of humans.
This is because the conventional method could only measure response values of robots and could not predict command values.
As a result, a response value of the next step is calculated as a command value.
This fact requires an assumption such that response and command values are identical with the existence of ideal controllers.

In order to solve this problem, the author's research group proposed bilateral control-based imitation learning generating motions without the assumption such that the command and response values are identical \cite{adachi18:_imitat_learn_objec_manip_based}\cite{fujimoto19:_time_series_motion_gener_consid}\cite{ayumu20:_imitat_learn_based_bilat_contr}\cite{sasagawa21:_motion_gener_using_bilat_contr}.
In the proposed method, the motion to the robot is demonstrated by using bilateral control, which is a kind of teleoperation \cite{kitamura16:_bilat_contr_vertic_direc_using}\cite{sakaino17:_bilat_contr_elect_hydraul_actuat}\cite{sakaino09:_obliq_coord_contr_advan_motion_contr}.
Bilateral control does not only synchronizes two robots (master and slave robots), but also transfers the reaction force on the teleoperated slave robot to the master robot.
Therefore, motions including human skills to manipulate environments can be taught.
Furthermore, because response values of the master robot are given as command values of the slave robot in bilateral control, both response and command values of the slave robot can be measured. 
As a result, it is possible to predict the slave robot's next-step command values corresponding to the next-step response values of the master robot enabling human-level high-speed operation.

However, bilateral control-based imitation learning requires robots that can control torque.
Then, it cannot be implemented in robots that can only follow position and velocity commands like many commercially available robots.
To overcome the problem, in this research, a way to implement bilateral control-based imitation learning to velocity-controlled robot is proposed.
In the proposed method, torque references are fist calculated, and they are transformed to velocity commands by using an admittance control-like procedure.
Even though the key idea is very straightforward, the superiority of the bilateral control-based imitation learning; fast and adaptive motions, maintains.
The effectiveness of the proposed method is evaluated by a mopping task.
By using the proposed method, robots could respond to fluctuations in the length of the mop handle in real-time, while at the same time achieving human-level fast movements.

This paper is organized as follows. 
The robot and controllers are explained in section~\ref{sec:robot}.
Then, section~\ref{sec:imitation_learning} describes the proposed imitation learning.
The validity of the proposed method is experimentally verified in section~\ref{sec:experiment}.
Finally, this study is concluded in section~\ref{sec:conclusion}.

%%%%%%%%%%%%%%%%%%%%%%%%%%%%%%%%%%%%%%%%%%%%%%%%%%%%%%%%%%%%
\section{Robot and controller} \label{sec:robot}
In this section, robots and controllers are described.
\subsection{Robot} 
\begin{figure}[t] %1 
  \centering 
  \includegraphics[width=40mm]{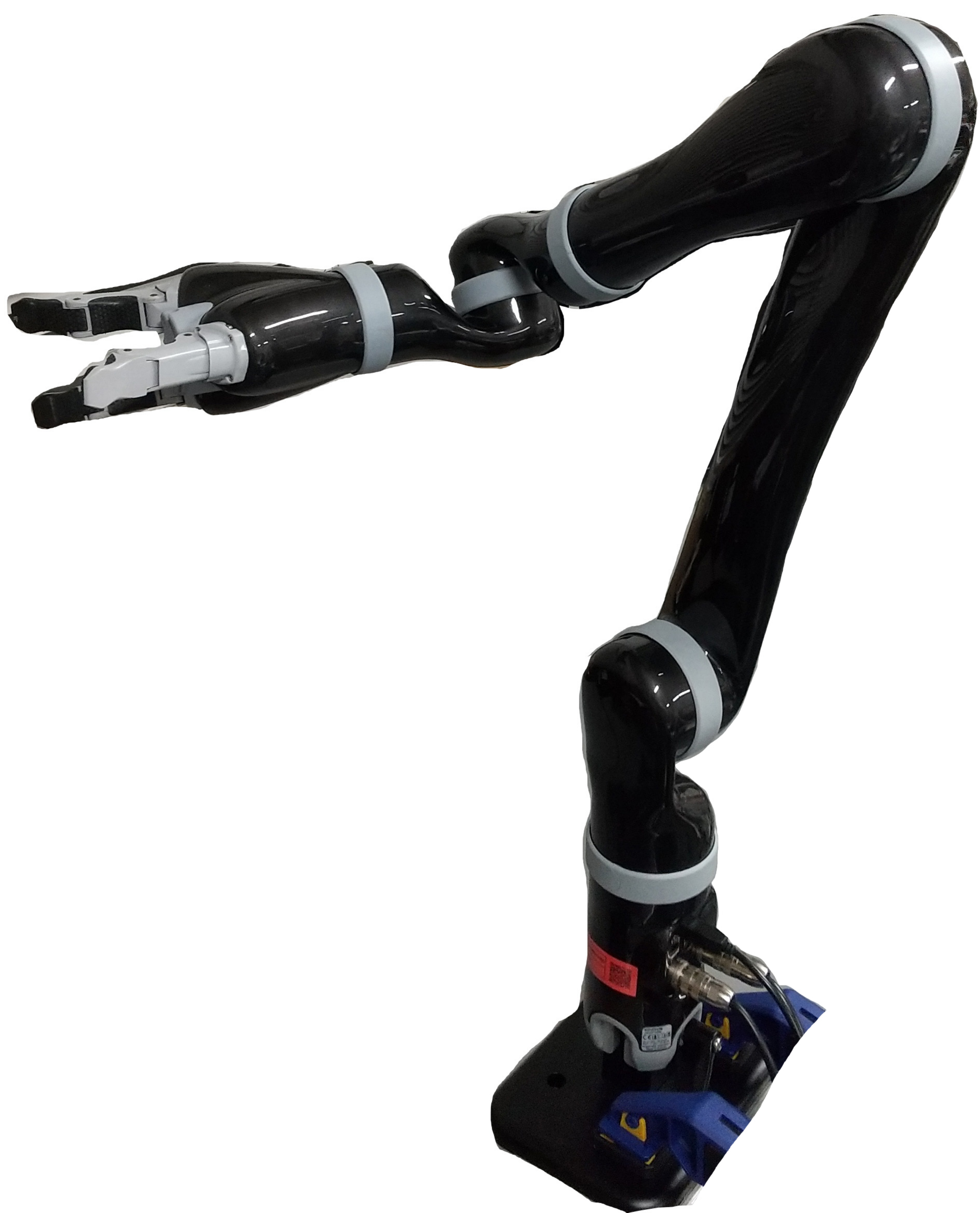}
  \caption{Robot}
  \label{jaco2}
\end{figure}
In this study, two Kinova Gen2 (6 DOF, KG-3) shown in Fig.~\ref{jaco2} were used as master and slave robots.
From the base to the tips of the robots, the joints were numbered 0, 1, ..., 5 in that order.
Angle responses of each joint were measured by optical encoders, angular velocity responses were obtained by using pseudo differential, and torque responses of each joint were measured by torque sensors, respectively.
The robots were set to a velocity control mode where the robots only track a desired velocity command.
In the controller designing, we assumed that an inertial matrix, $\ve{J}$, of the robots were constant diagonal matrices such as
\begin{eqnarray}
\ve{J} = \rm{diag}\it{[J_0, J_1, J_2, J_3, J_4, J_5]}
\end{eqnarray}
where $J_i$ is the inertia around the \textit{i}-th axis.
The physical parameters of the robot were identified by extending the method given in \cite{yamazaki17:_estim_kinet_model_human_arm}, and the parameters are presented in TABLE~\ref{tbl:id_parameter}.
\begin{table}
\centering
\caption{Identified system parameters}
\label{tbl:id_parameter}
\begin{tabular}{@{}l|ccc}
\hline
      $J_{0}$ & Joint~1's inertia [{\rm m Nm}] & 0.939 \\
      $J_{1}$ & Joint~2's inertia [{\rm m Nm}] & 1.32 \\
      $J_{2}$ & Joint~3's inertia [{\rm m Nm}] & 1.32 \\
      $J_{3}$ & Joint~3's inertia [{\rm m Nm}] & 0.363 \\
      $J_{4}$ & Joint~3's inertia [{\rm m Nm}] & 0.196\\
      $J_{5}$ & Joint~3's inertia [{\rm m Nm}] & 0.246\\
\hline
\end{tabular}
\end{table}

\subsection{Four-Channel Bilateral Control}
\begin{figure}[t] %8
  \centering 
  \includegraphics[width=90mm]{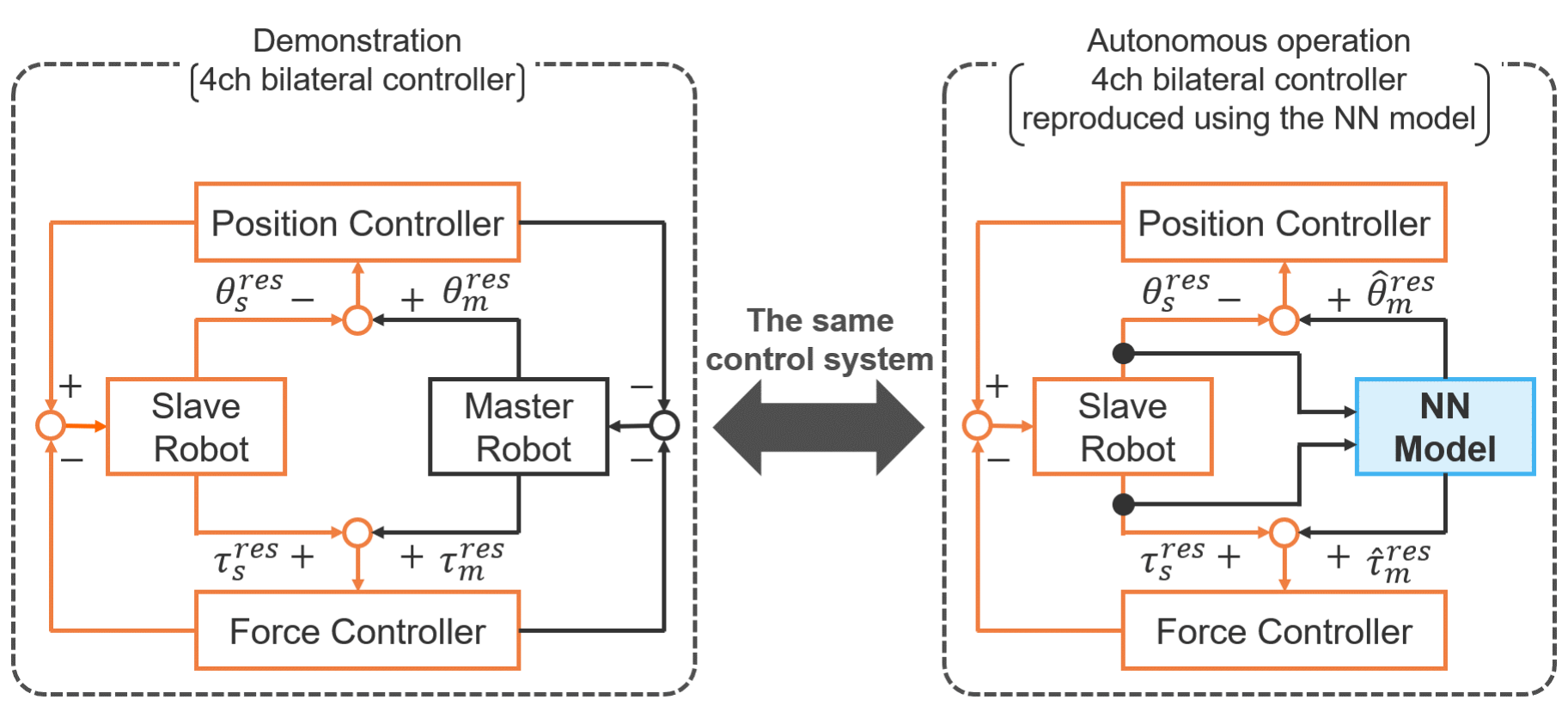}
  \caption{
Four-channel bilateral controller, in the demonstrations (left) and in the autonomous operations (right).
}
  \label{4ch}
\end{figure}

In this study, acceleration control was realized by using a disturbance observer (DOB) to reject disturbances on the robots \cite{DOB}.
The overview of four-channel bilateral control is shown in the left figure of 
Fig.~\ref{eq:4ch_controller}.
Then, four-channel bilateral control was implemented \cite{4ch}.
Here, a position controller was composed of a proportional and differential controller, and a force controller was composed of a proportional controller, respectively.
Variables $\theta$, $\dot{\theta}$, and $\tau$ represent the joint angle, angular velocity, and torque, respectively.
Variables with the superscripts $cmd$, $res$, and $ref$ mean the command, response, and reference values, respectively.
In torque-controlled robots, torque references of the master and slave controllers,  $\ve{\tau}_m^{ref}$ and $\ve{\tau}_s^{ref}$, are given as follows:
\begin{eqnarray}
\label{eq:4ch_controller}
\ve{\tau}_m^{ref}= \ve{J}(\ve{K_p}+\ve{K_d} s)(\ve{\theta}_s^{res}-\ve{\theta}_m^{res})-\ve{K_f}(\ve{\tau}^{res}_m+\ve{\tau}^{res}_s)\\
\ve{\tau}_s^{ref}=  \ve{J}(\ve{K_p}+\ve{K_d} s)(\ve{\theta}_m^{res}-\ve{\theta}_s^{res})-\ve{K_f}(\ve{\tau}^{res}_m+\ve{\tau}^{res}_s)
\end{eqnarray}
where 
\begin{eqnarray}
\ve{K_p}&=&\rm{diag}\it[K_{p1},K_{p2},K_{p3},K_{p4},K_{p5},K_{p6}]\\\nonumber \ve{K_d}&=&\rm{diag}\it[K_{d1},K_{d2},K_{d3},K_{d4},K_{d5},K_{d6}]\\\nonumber \ve{K_f}&=&\rm{diag}\it[K_{f1},K_{f2},K_{f3},K_{f4},K_{f5},K_{f6}]\nonumber
\end{eqnarray}
are control gains, and $s$ is the Laplace operator, respectively.
The parameters $K_{pi}$, $K_{vi}$, and $K_{fi}$ are feedback gains of the \textit{i}-th joint.
Variables $\ve{\theta}_m$, $\ve{\theta}_s$, $\ve{\tau}_m$, and $\ve{\tau}_s$ describe the slave states as follows:
\begin{eqnarray}
 \ve{\theta_m}=
\left[
  \begin{array}{c}
  \theta_{m0} \\
  \theta_{m1} \\
  \theta_{m2} \\
  \theta_{m3} \\
  \theta_{m4} \\
  \theta_{m5} \\
  \end{array}
  \right]
&,&
 \ve{\theta_s}=
\left[
  \begin{array}{c}
  \theta_{s0} \\
  \theta_{s1} \\
  \theta_{s2} \\
  \theta_{s3} \\
  \theta_{s4} \\
  \theta_{s5} \\
  \end{array}
  \right]
,\nonumber\\
 \ve{\tau_m}=
\left[
  \begin{array}{c}
  \tau_{m0} \\
  \tau_{m1} \\
  \tau_{m2} \\
  \tau_{m3} \\
  \tau_{m4} \\
  \tau_{m5} \\
  \end{array}
  \right]
  &,&
 \ve{\tau_s}=
\left[
  \begin{array}{c}
  \tau_{s0} \\
  \tau_{s1} \\
  \tau_{s2} \\
  \tau_{s3} \\
  \tau_{s4} \\
  \tau_{s5} \\
  \end{array}
  \right]
\end{eqnarray}
where the subscripts $mi$ and $si$ indicate the \textit{i}-th joint of the master and slave robots.

\subsection{Velocity Control}
However, these controllers require to control torque of joints.
It is impossible for many industrial robots that can only track position or velocity commands.
Therefore, in this study, we proposed to implement an admittance control-like procedure to obtain velocity commands.
Then, angular velocity commands should satisfy the following relations with virtual mass and damping coefficients, $\ve{M}=\rm{diag}\it[M_0, M_1, M_2, M_3, M_4, M_5]$ and $\ve{D}$.
\begin{eqnarray}
 \ve{M}\ddve{\theta}_m^{cmd} + \ve{D} \dve{\theta}_m^{cmd} &=& \ve{\tau}_m^{ref}\\
 \ve{M}\ddve{\theta}_s^{cmd} + \ve{D} \dve{\theta}_s^{cmd} &=& \ve{\tau}_s^{ref}
\end{eqnarray}
Then, angular velocity commands are obtained as follows:
\begin{eqnarray}
\label{eq:dth_cmd_m}
 \dve{\theta}_m^{cmd} &=& \frac{1}{\ve{M}s+\ve{D}}\ve{\tau}_m^{ref}
 =\frac{1}{\ve{M}}\frac{1}{s+\omega}\ve{\tau}_m^{ref}\\
 \label{eq:dth_cmd_s}
 \dve{\theta}_s^{cmd} &=& \frac{1}{\ve{M}s+\ve{D}}\ve{\tau}_s^{ref}
  =\frac{1}{\ve{M}}\frac{1}{s+\omega}\ve{\tau}_s^{ref}
\end{eqnarray}
where $\omega$ is a cut-off frequency, and the damping coefficient was determined as $\ve{D}=\omega\ve{M}$.
Then, finally four-channel bilateral control can be implemented in velocity-controlled robots.
Control parameters used in this study are shown in TABLE~\ref{tbl:gain}.
\begin{table}
\centering
\caption{Gains of robot controller}
\label{tbl:gain}
\begin{tabular}{c|ccc}
 \hline
      $K_{p0}$ & Position feedback gain & 9 \\
      $K_{p1}$ & Position feedback gain & 16 \\
      $K_{p2}$ & Position feedback gain & 16 \\
      $K_{p3}$ & Position feedback gain & 4 \\
      $K_{p4}$ & Position feedback gain & 9 \\
      $K_{p5}$ & Position feedback gain & 16 \\
      $K_{d0}$ & Velocity feedback gain & 6 \\
      $K_{d1}$ & Velocity feedback gain & 8 \\
      $K_{d2}$ & Velocity feedback gain & 8 \\
      $K_{d3}$ & Velocity feedback gain & 4 \\
      $K_{d4}$ & Velocity feedback gain & 6 \\
      $K_{d5}$ & Velocity feedback gain & 8 \\
      $K_{f0}$ & Force feedback gain & 0.13 \\
      $K_{f1}$ & Force feedback gain & 0.05 \\
      $K_{f2}$ & Force feedback gain & 0.05 \\
      $K_{f3}$ & Force feedback gain & 0.10 \\
      $K_{f4}$ & Force feedback gain & 0.2 \\
      $K_{f5}$ & Force feedback gain & 0.2 \\
      $M_{0}$ & Virtual mass & 0.2 \\
      $M_{1}$ & Virtual mass & 0.5 \\
      $M_{2}$ & Virtual mass & 0.5 \\
      $M_{3}$ & Virtual mass & 0.3 \\
      $M_{4}$ & Virtual mass & 0.1 \\
      $M_{5}$ & Virtual mass & 0.1 \\
      $\omega$ & Cut-off frequency of admittance control [{\rm rad/s}] & 30.0 \\ 
      $g$ & Cut-off frequency of pseudo derivative [{\rm rad/s}] & 20.0 \\ 
      $g_{DOB}$ & Cut-off frequency of DOB [{\rm rad/s}] & 10.0 \\ 
 \hline
\end{tabular}
\end{table}

\section{Imitation Learning}
\label{sec:imitation_learning}
Here, bilateral control-based imitation learning for velocity-controlled robots is explained.
\subsection{Data collection phase}
In bilateral control-based imitation learning, teacher data was collected using four-channel bilateral control.
Note that the controller used here is given by (\ref{eq:dth_cmd_m}) and (\ref{eq:dth_cmd_s}).

\begin{figure}[t] %1 
  \centering 
  \includegraphics[width=40mm]{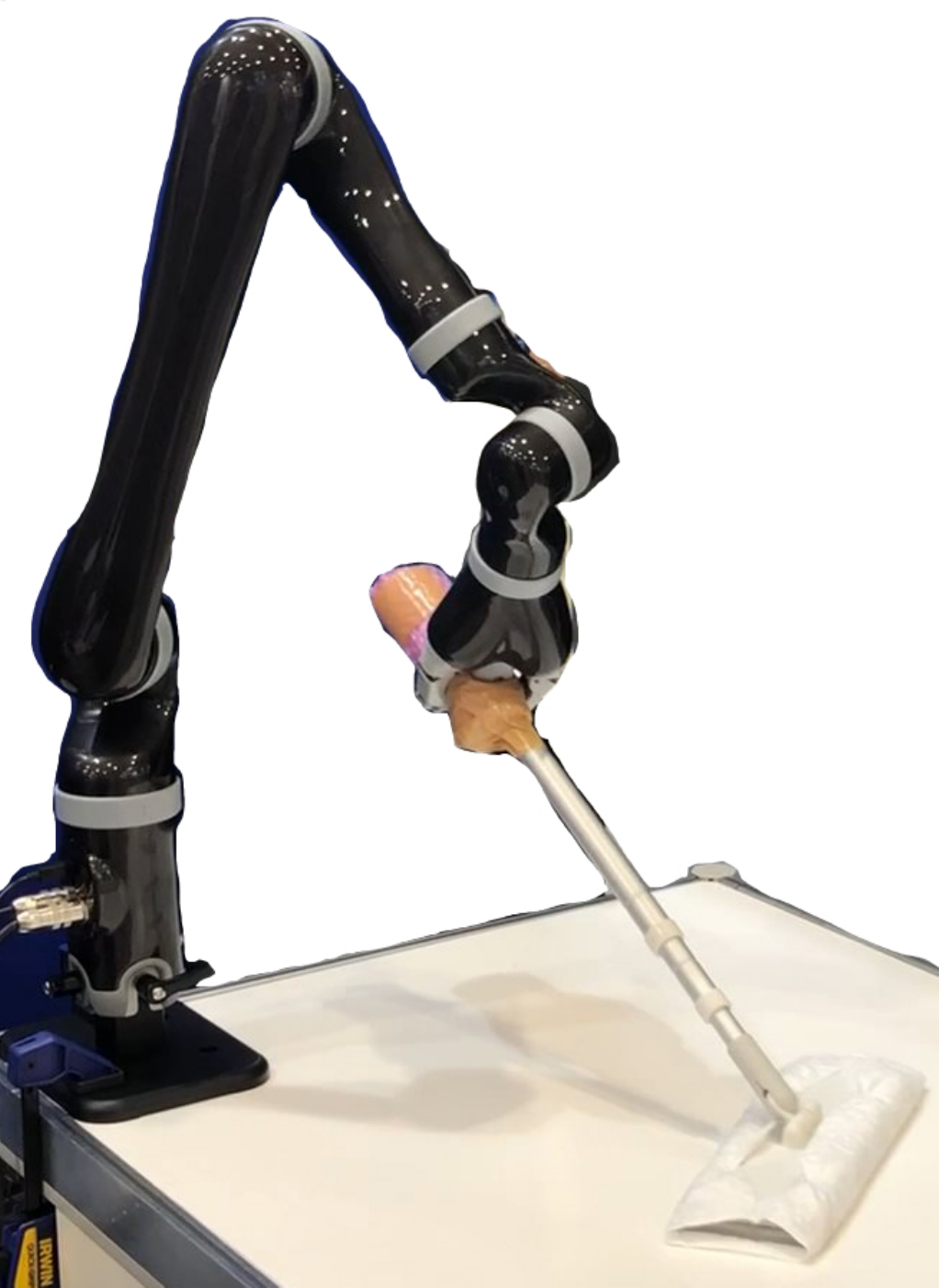}
  \caption{Mopping Task}
  \label{mopping}
\end{figure}
In this study, a mopping task was achieved.
Fig.~\ref{mopping} is a schematic of the task.
The mop handle was too thin to be grasped by the robot's hand, so a plastic cushion was wrapped around the mop handle. We also made the robot grab the mop by opening and closing the hand with a manual remote controller.
Then, the robot wiped a surface of a desk.
15 trials were conducted with different length of the mop handle.
Then, the responses of the master and slave were collected.

\subsection{Training phase }
\begin{figure}[t] %1 
  \centering 
  \includegraphics[width=80mm]{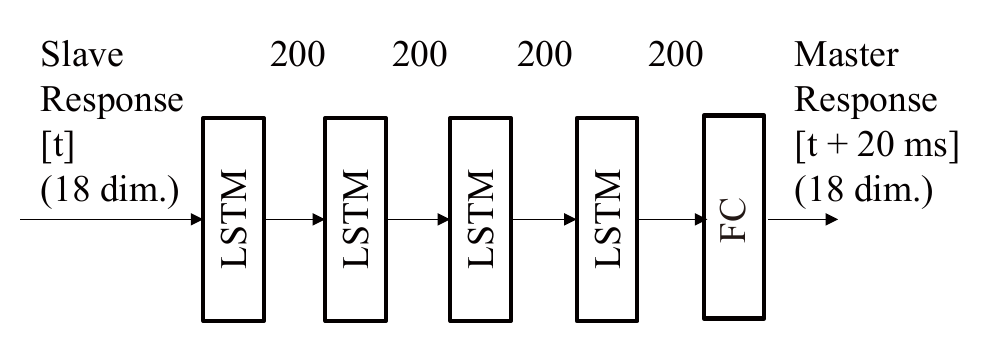}
  \caption{Neural network model}
  \label{NN_model}
\end{figure}
In this study, a neural network (NN) was trained to predict master's responses from slave's responses.
Recurrent neural networks are suitable for inference of time series data \cite{zhang16:_highw_rnns}\cite{zhang17:_three_recur_neural_networ_three}.
Especially, a long short-term memory (LSTM) model \cite{hochreiter97:_long_short_term_memor} was used to achieve long and short time series inference.
As shown in Fig.~\ref{NN_model}, the NN had four LSTM layers with 200 units and a fully connected (FC) layer.
The inputs were 18 dimensional slave's responses (6 [axes] $\times$ 3 [variables]: $\ve{\theta}_s^{res}, \dve{\theta}_s^{res}$, and $\ve{\tau}_s^{res}$).
The outputs were the next master's responses: $\hat{\ve{\theta}}_m^{res}$, $\hat{\dve{\theta}}_m^{res}$, and $\hat{\ve{\tau}}_m^{res}$.
Because the NN was updated in 20 ms, the predicted master response was 20 ms after the slave response.
Then the training data was 20 times augmented by using the technique shown in \cite{rahmatizadeh16:_from_virtual_demon_real_world}.
Input values were normalized by using min-max normalization.
The 100 time series batch was extracted as a mini-batch in the learning of the NN.

\subsection{Autonomous operation phase}
In the autonomous operation phase, estimates of master responses were obtained by using the trained NN model.
In other word, the NN model substituted for the slave as shown in the right figure of Fig.~\ref{eq:4ch_controller}.
Note that the controller in the slave side was not changed from the data collection as follows:
\begin{eqnarray}
\label{eq:autonomous_controller}
\ve{\tau}_s^{ref}=  \ve{J}(\ve{K_p}+\ve{K_d} s)(\ve{\hat{\theta}}_m^{res}-\ve{\theta}_s^{res})-\ve{K_f}(\ve{\hat{\tau}}^{res}_m+\ve{\tau}^{res}_s)
\end{eqnarray}
\begin{eqnarray}
 \dve{\theta}_s^{cmd} 
  =\frac{1}{\ve{M}}\frac{1}{s+\omega}\ve{\tau}_s^{ref}.
\end{eqnarray}
If we change the controllers, the skills of the operators to teleoperate slave robots are lost.
However, our bilateral control-based imitation learning can maintain the controller, and this is the main contribution of our approach.

\begin{figure*}[t] %1 
  \centering 
  \includegraphics[width=140mm]{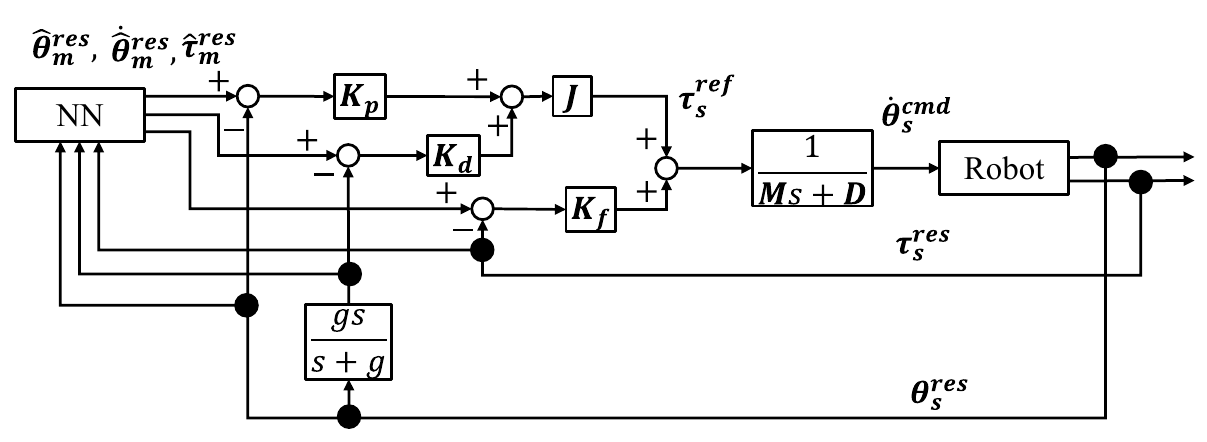}
  \caption{Block diagram of autonomous operation}
  \label{block}
\end{figure*}

Fig.~\ref{block} shows the entire block diagram in the autonomous operation phase.

\section{Experiment}
\label{sec:experiment}
In this section, the proposed method is experimentally evaluated.
In this experiments, we evaluated whether the robot could correctly  mop even when the length of the mop handle was changed.
First, experiments with two different mop length were conducted where two lengths from the tip to the grasping point were 480 mm and 530 mm.

\begin{figure}[t] %1 
  \centering 
  \includegraphics[width=80mm]{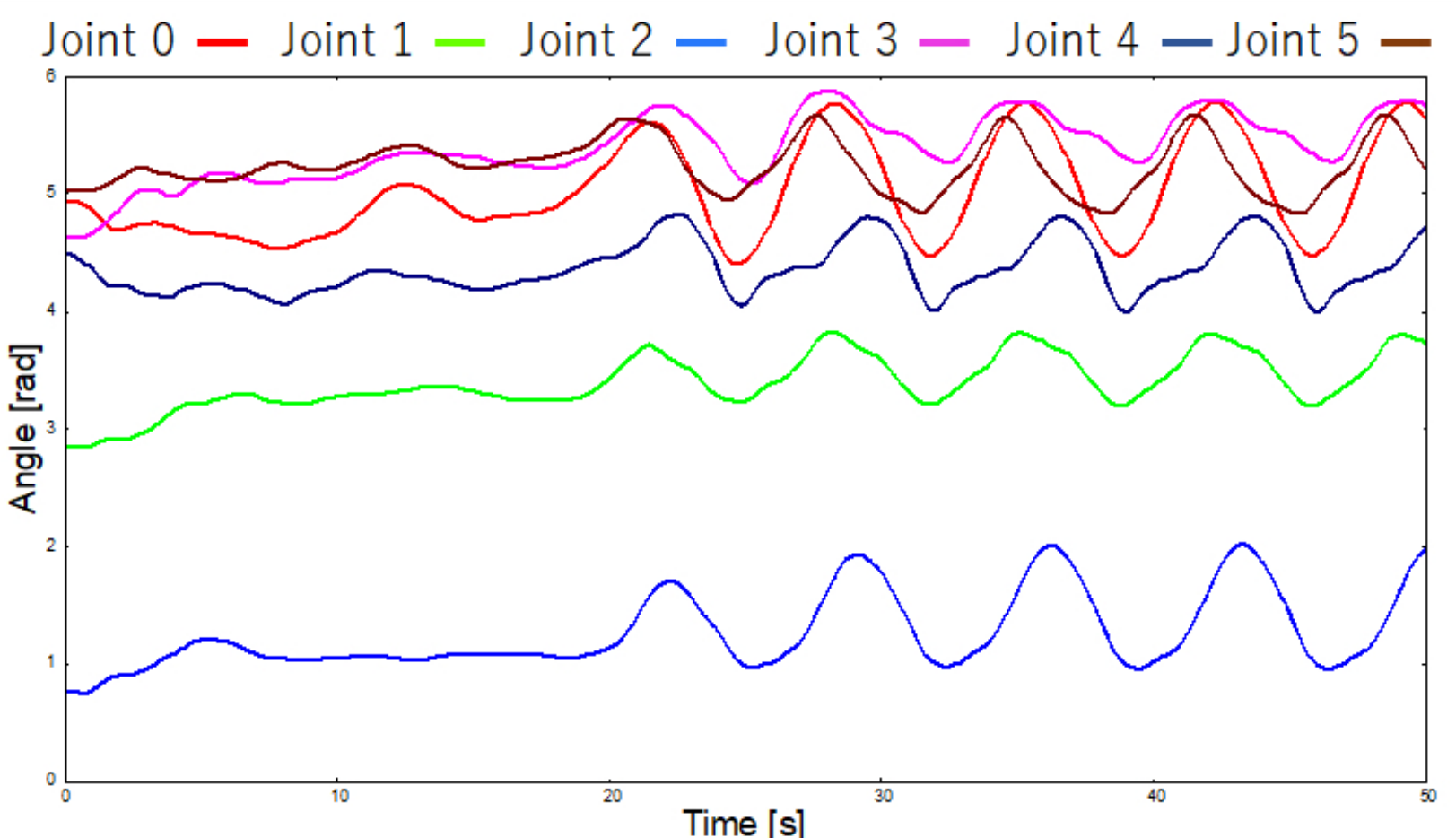}
  \caption{Experimental result (angle response, 480mm)}
  \label{x_480}
\end{figure}

\begin{figure}[t] %1 
  \centering 
  \includegraphics[width=80mm]{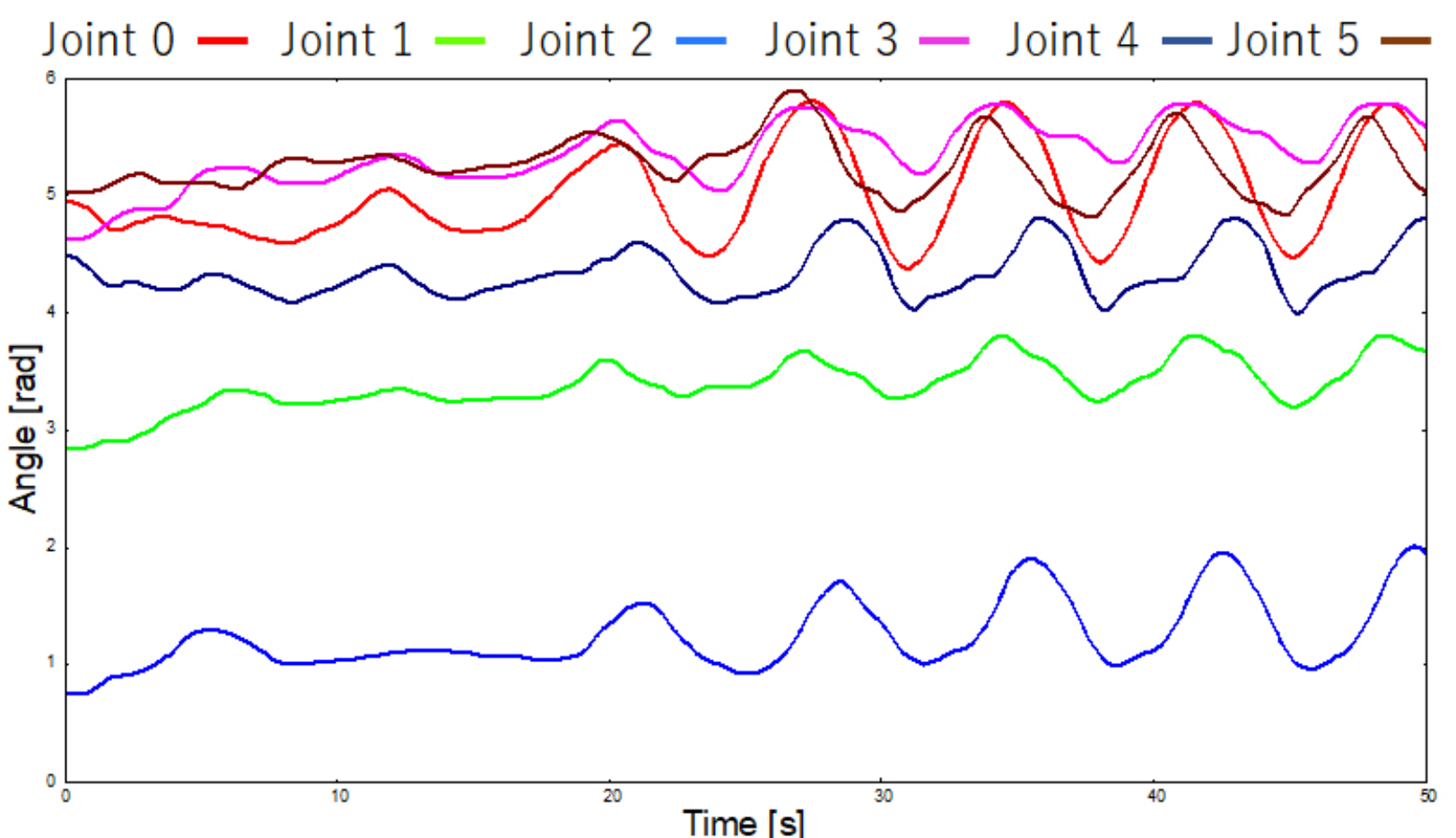}
  \caption{Experimental result (angle response, 530mm)}
  \label{x_530}
\end{figure}

\begin{figure}[t] %1 
  \centering 
  \includegraphics[width=80mm]{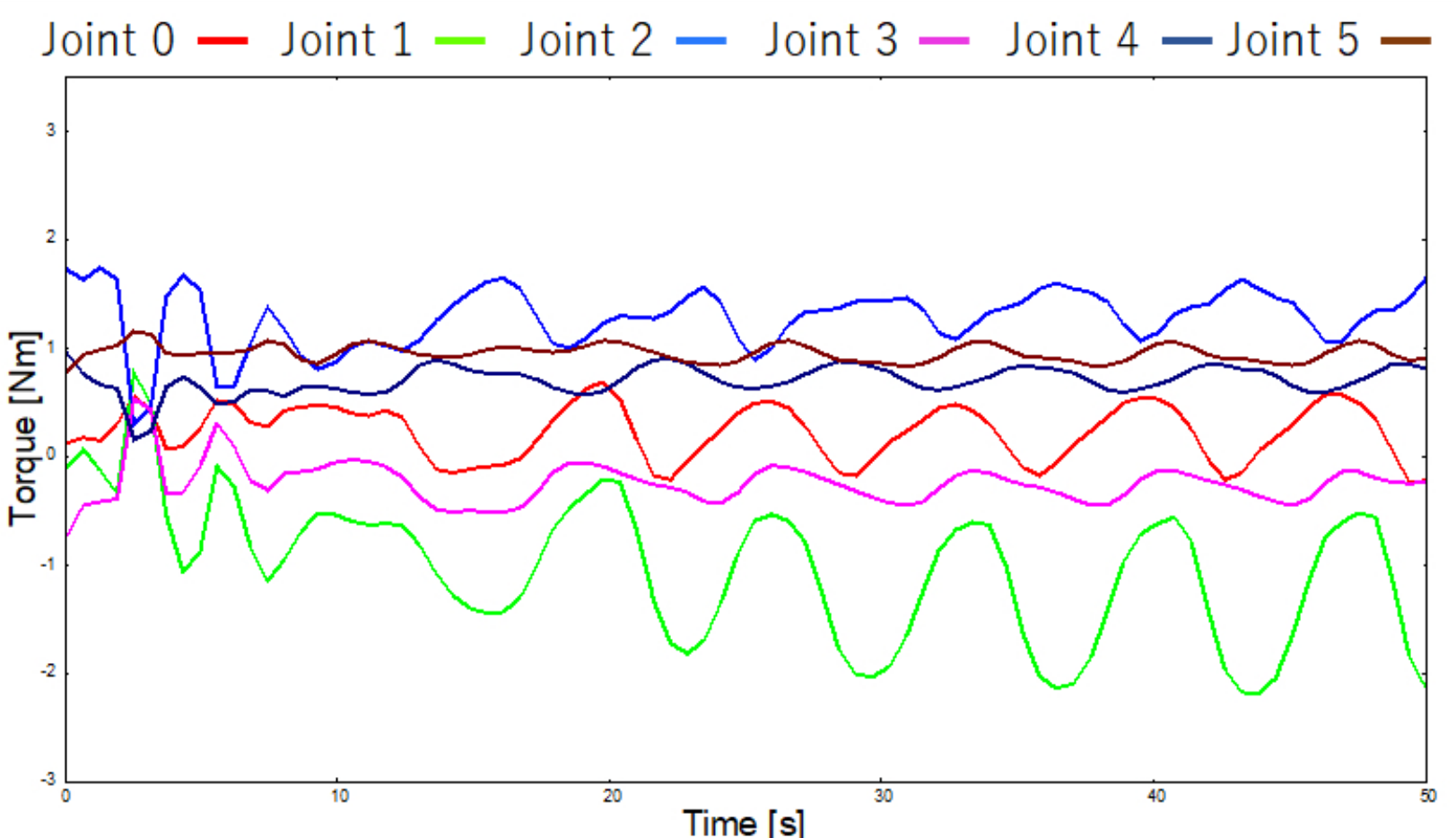}
  \caption{Experimental result (torque response, 480mm)}
  \label{f_480}
\end{figure}

\begin{figure}[t] %1 
  \centering 
  \includegraphics[width=80mm]{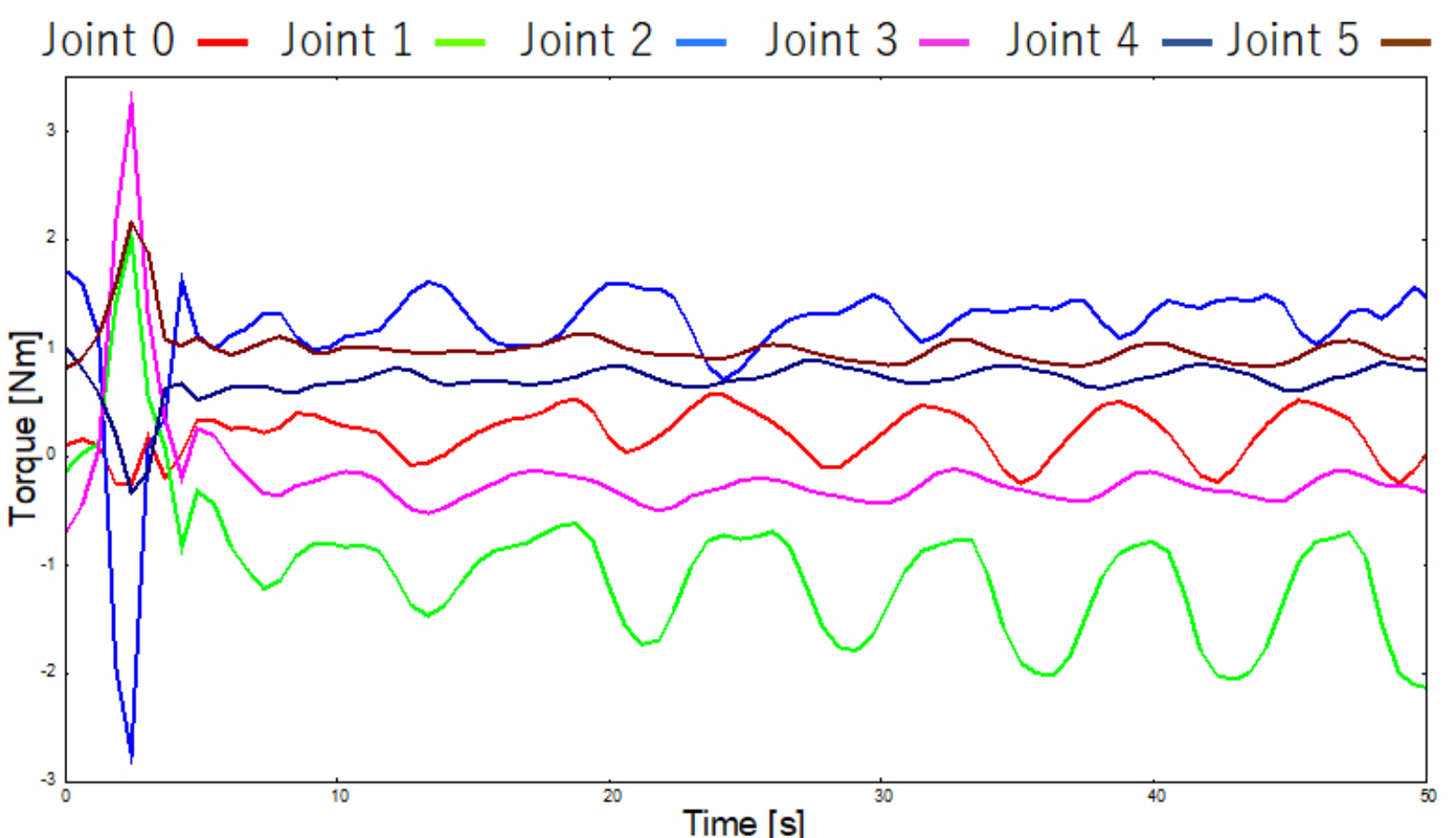}
  \caption{Experimental result (torque response, 530mm)}
  \label{f_530}
\end{figure}

Figs.~\ref{x_480} and \ref{x_530} are angle responses, and Figs.~\ref{f_480} and \ref{f_530} are torque responses, respectively.
The first 20 seconds were spent for adjusting to the mopping posture.
After that, the robot mopped four times.
As we can see from Figs.~\ref{x_480} and \ref{x_530}, the difference in each length was very small in angle responses.
On the contrary, we can find the difference in torque responses as shown in Figs.~\ref{f_480} and \ref{f_530}.
In the case of 530 mm, the mop was so long that it collided with the desk in the initializing to the mopping posture, resulting in an irregular response, but even in such a case, an appropriate torque command value could be generated without accumulation of errors, and mopping could be performed.
In addition, looking at the response of joint 1, which corresponds to the joint to extend the arm, it can be seen that the amplitude of the torque was larger at 480 mm than 530 mm.
This is consistent with the fact that the mop was shorter and required more torque to maintain contact with the desk.

When robust control is used, such environmental fluctuations will be suppressed as disturbances and the system will be forced to follow a command value.
In the present situation, it is likely that the position tracking performance will be excessively enhanced and excessive contact force will be generated, or the force tracking performance will be excessively enhanced and the position will not track to desired positions at all. 
It would be very difficult to realize this task within the scope of conventional robust control because the appropriate impedance changes with time in this task.
For this reason, behavioral cloning like the motion copying system \cite{yokokura08:_motion_copyin_system_real_haptic_variab_speed}\cite{chinthaka18:_inert_compen_motion_copyin_system} is not effective when there are environmental changes.
However, the proposed method can probabilistically estimate appropriate command values under environmental fluctuations that could be assumed from multiple past actions, and thus could continue to perform the task even with the fluctuation of tools and environments.
Because the proposed method has a mechanism for estimating both the position and force commands, it is equivalently possible to realize time-varying impedance control.


\begin{figure*}[t] %1 
  \centering 
  \includegraphics[width=130mm]{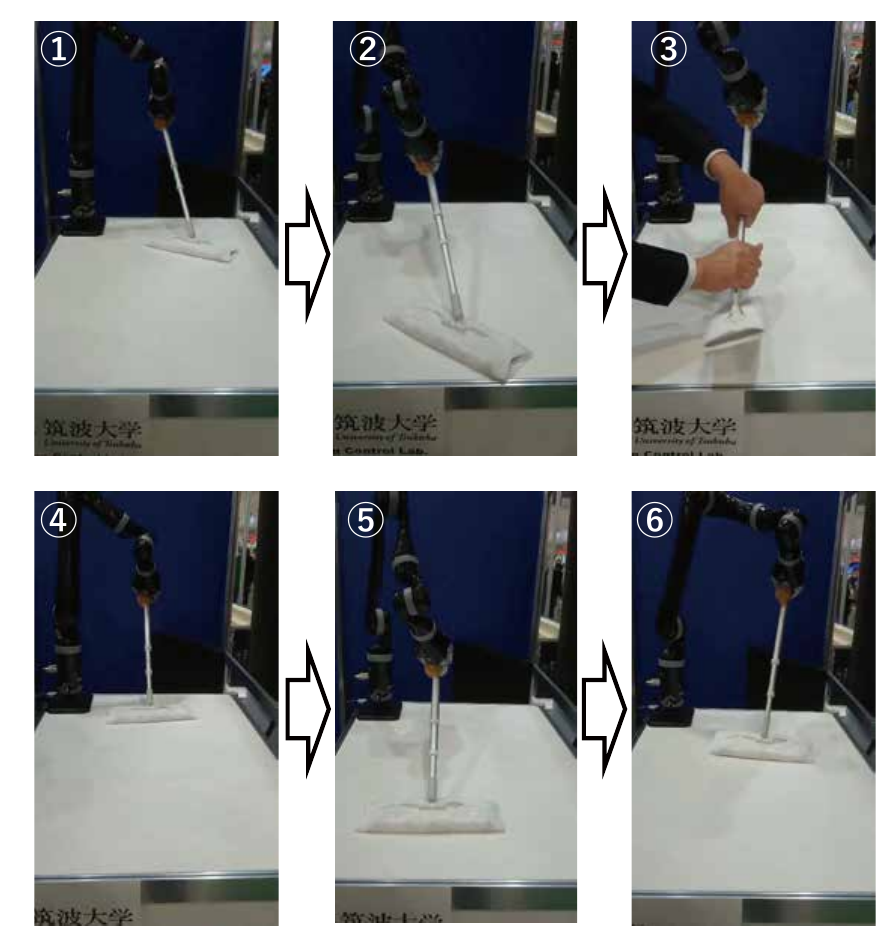}
  \caption{Snapshot of experiment to change the length}
  \label{snapshot}
\end{figure*}

In order to further demonstrate the usefulness of the proposed method, the author demonstrated to change the length of the mop in mopping.
Fig.~\ref{snapshot} shows a snapshot of the experiment.
Even though the length was changed during the operation, the robot could keep mopping.
\section{Conclusion}
\label{sec:conclusion}
In this study, a way to implement bilateral control-based imitation learning in velocity-controlled robots was proposed.
The torque references were transformed into velocity commands by using an admittance control-like procedure.
Although the proposed procedure is very straightforward, high adaptability for environmental changes and movements at human-level high-speed were achieved at the same time.
Note that the proposed procedure can be easily extended to position-controlled robots by using admittance control.
But of course, there is a drawback when compared to the method implemented on robots that can control torque.
That is, because the proposed method has the form of admittance control, the control performance cannot be made sufficiently high because the torque reference is integrated.
As a result, the operation feeling of the bilateral control was very heavy, and it was difficult to teach quick and delicate motions such as turning over pancakes \cite{Tsuji2016}\cite{tsuji18:_optim_trajec_gener_model_predic}.

\bibliographystyle{ieeetr}
\bibliography{IEEEabrv,../ref/ref}

\begin{thebibliography}{10}

\bibitem{levine17:_learn}
S.~Levine, P.~Pastor, A.~Krizhevsky, J.~Ibarz, and D.~Quillen, ``Learning
  hand-eye coordination for robotic grasping with deep learning and large-scale
  data collection,'' {\em The International Journal of Robotics Research},
  vol.~37, no.~4-5, pp.~421--436, 2017.

\bibitem{tsurumine19:_deep}
Y.~Tsurumine, Y.~Cui, E.~Uchibe, and T.~Matsubara, ``Deep reinforcement
  learning with smooth policy update: Application to robotic cloth
  manipulation,'' {\em Robotics and Autonomous Systems}, vol.~112, pp.~72--83,
  2019.

\bibitem{yang16:_repeat_foldin_task_human_robot}
P.-C. Yang, K.~Sasaki, K.~Suzuki, K.~Kase, S.~Sugano, and T.~Ogata,
  ``Repeatable folding task by humanoid robot worker using deep learning,''
  {\em IEEE Robotics and Automation Letters}, vol.~2, no.~2, pp.~397--403,
  2016.

\bibitem{Calinon2007}
S.~Calinon, F.~Guenter, and A.~Billard, ``{On learning, representing, and
  generalizing a task in a humanoid robot},'' {\em IEEE Transactions on
  Systems, Man, and Cybernetics, Part B: Cybernetics}, vol.~37, no.~2,
  pp.~286--298, 2007.

\bibitem{kyrarini19:_robot_learn_indus_assem_task_human_demon}
M.~Kyrarini, M.~A. Haseeb, D.~Ristic-Durrant, and A.~G.~P. Graeser, ``Robot
  learning of industrial assembly task via human demonstrations,'' {\em
  Autonomous Robots}, vol.~43, pp.~239--257, 2019.

\bibitem{kormushev11:_imitat_learn_posit_force_skill}
P.~Kormushev, S.~Calinon, and D.~G. Caldwell, ``Imitation learning of
  positional and force skills demonstrated via kinesthetic teaching and haptic
  input,'' {\em Advanced Robotics}, vol.~25, no.~5, pp.~581--603, 2011.

\bibitem{zhang18:_deep_imitat_learn_compl_manip}
T.~Zhang, Z.~McCarthy, O.~Jow, D.~Lee, X.~Chen, K.~Goldberg, and P.~Abbeel,
  ``Deep imitation learning for complex manipulation tasks from virtual reality
  teleoperation,'' in {\em Proceedings of 2018 IEEE International Conference on
  Robotics and Automation (ICRA)}, pp.~5628--5635, 2018.

\bibitem{jiny20:_geomet_persp_visual_imitat_learn}
J.~Jiny, L.~Petrichy, M.~Dehghany, and M.~Jagersand, ``A geometric perspective
  on visual imitation learning,'' in {\em Proceedings of the INternational
  Conference on Intelligent Robots and Systems}, pp.~5194--5200, 2020.

\bibitem{lee15:_learn_force_based_manip_defor}
A.~X. Lee, H.~Lu, A.~Gupta, S.~Levine, and P.~Abbeel, ``Learning force-based
  manipulation of deformable objects from multiple demonstrations,'' in {\em
  2015 IEEE International Conference on Robotics and Automation}, pp.~177--184,
  2015.

\bibitem{rozo13}
L.~Rozo, P.~Jimenez, and C.~Torras, ``A robot learning from demonstration
  framework to perform force-based manipulation tasks,'' {\em Intel Serv
  Robotics}, vol.~6, pp.~33--51, 2013.

\bibitem{ochi18:_deep_learn_scoop_motion_using_bilat_teleop}
H.~Ochi, W.~Wan, Y.~Yang, N.~Yamanobe, J.~Pan, and K.~Harada, ``Deep learning
  scooping motion using bilateral teleoperations,'' in {\em Proceedings of 2018
  3rd International Conference on Advanced Robotics and Mechatronics (ICARM)},
  pp.~118--123, 2018.

\bibitem{rozo19:_inter_trajec_adapt_force_baysian_optim}
L.~Rozo, ``Interactive trajectory adaptation through force-guided baysian
  optimization,'' in {\em Proceedings of 2019 IEEE/RSJ International Conference
  on Intelligent Robots and Systems (IROS)}, p.~7603, 2019.

\bibitem{osa18:_onlin_trajec_plann_force_contr}
T.~Osa, N.~Sugita, and M.~Mitsuishi, ``Online trajectory planning and force
  control for automation of surgical tasks,'' {\em IEEE Transactions on
  Automation Science and Engineering}, vol.~15, no.~2, pp.~675--691, 2018.

\bibitem{rozo15:_learn}
L.~Rozo, D.~Bruno, S.~Calinon, and D.~G. Caldwell, ``{Learning optimal
  controllers in human-robot cooperative transportation tasks with position and
  force constraints},'' in {\em Proceedings of IEEE/RSJ International
  Conference on Intelligent Robots and Systems}, pp.~1024--1030, 2015.

\bibitem{gupta2019relay}
A.~Gupta, V.~Kumar, C.~Lynch, S.~Levine, and K.~Hausman, ``Relay policy
  learning: Solving long-horizon tasks via imitation and reinforcement
  learning,'' {\em arXiv:1910.11956}, 2019.

\bibitem{adachi18:_imitat_learn_objec_manip_based}
T.~Adachi, K.~Fujimoto, S.~Sakaino, and T.~Tsuji, ``Imitation learning for
  object manipulation based on position/force information using bilateral
  control,'' in {\em Proceedings of the 2018 IEEE/RSJ International Conference
  on Intelligent Robots and Systems}, pp.~3648--3653, 2018.

\bibitem{fujimoto19:_time_series_motion_gener_consid}
K.~Fujimoto, S.~Sakaino, and T.~Tsuji, ``Time series motion generation
  considering long short-term moiton,'' in {\em Proceedings of 2019 IEEE/RSJ
  International Conference on Intelligent Robots and Systems (IROS)},
  pp.~6842--6848, 2019.

\bibitem{ayumu20:_imitat_learn_based_bilat_contr}
A.~Sasagawa, K.~Fujimoto, S.~Sakaino, and T.~Tsuji, ``Imitation learning based
  on bilateral control for human-robot cooperation,'' {\em IEEE Robotics and
  Automation Letters}, vol.~5, no.~4, pp.~6169--6176, 2020.

\bibitem{sasagawa21:_motion_gener_using_bilat_contr}
A.~Sasagawa, S.~Sakaino, and T.~Tsuji, ``Motion generation using bilateral
  control-based imitation learning with autoregressive learning,'' {\em IEEE
  Access}, vol.~9, pp.~20508--20520, 2021.

\bibitem{kitamura16:_bilat_contr_vertic_direc_using}
T.~Kitamura, N.~Mizukami, H.~Mizoguchi, S.~Sakaino, and T.~Tsuji, ``Bilateral
  control in the vertical direction using functional electrical stimulation,''
  {\em IEEJ Journal of Industry Applications}, vol.~5, no.~5, pp.~398--404,
  2016.

\bibitem{sakaino17:_bilat_contr_elect_hydraul_actuat}
S.~Sakaino, T.~Furuya, and T.~Tsuji, ``Bilateral control between electric and
  hydraulic actuators using linearization of hydraulic actuators,'' {\em IEEE
  Transactions on Industrial Electronics}, 2017.

\bibitem{sakaino09:_obliq_coord_contr_advan_motion_contr}
S.~Sakaino, T.~Sato, and K.~Ohnishi, ``Oblique coordinate control for advanced
  motion control -- applied to micro-macro bilateral control --,'' in {\em
  Proceedings of the 2009 IEEE International Conference on Mechatronics}, April
  2009.

\bibitem{yamazaki17:_estim_kinet_model_human_arm}
T.~Yamazaki, S.~Sakaino, and T.~Tsuji, ``Estimation and kinetic modeling of
  human arm using wearable robot arm,,'' {\em Electrical Enginnering in Japan},
  vol.~199, no.~3, pp.~57--67, 2017.

\bibitem{DOB}
K.~Ohnishi, M.~Shibata, and T.~Murakami, ``Motion control for advanced
  mechatronics,'' {\em IEEE/ASME Transactions on Mechatronics}, vol.~1, no.~1,
  pp.~56--67, 1996.

\bibitem{4ch}
Y.~Matsumoto, S.~Katsura, and K.~Ohnishi, ``An analysis and design of bilateral
  control based on disturbance observer,'' in {\em Proceedings of the 10th IEEE
  International Conference on Industrial Technology}, pp.~802--807, 2003.

\bibitem{zhang16:_highw_rnns}
Y.~Zhang, G.~Chen, D.~Yu, K.~Yaco, S.~Khudanpur, and J.~Glass, ``Highway long
  short-term memory rnns for distant speech recognition,'' in {\em Proceedings
  of 2016 IEEE International Conference on Acoustics, Speech and Signal
  Processing}, pp.~5755--5759, 2016.

\bibitem{zhang17:_three_recur_neural_networ_three}
Z.~Zhang, L.~Zheng, J.~Yu, Y.~Li, and Z.~Yu, ``Three recurrent neural networks
  and three numerical methods for solving a repetitive motion planning scheme
  of redundant robot manipulators,'' {\em IEEE/ASME Transactions on
  Mechatronics}, vol.~22, no.~3, pp.~1423--1434, 2017.

\bibitem{hochreiter97:_long_short_term_memor}
S.~Hochreiter and J.~Schmidhuber, ``Long short-term memory,'' {\em Neural
  Computation}, vol.~9, no.~8, pp.~1735--1780, 1997.

\bibitem{rahmatizadeh16:_from_virtual_demon_real_world}
R.~Rahmatizadeh, P.~Abolghasemi, A.~Behal, and L.~B\"{o}l\"{o}ni, ``From
  virtual demonstration to real-world manipulation using lstm and mdn,'' {\em
  arXiv:1603.03833}, 2016.

\bibitem{yokokura08:_motion_copyin_system_real_haptic_variab_speed}
Y.~Yokokura, S.~Katsura, and K.~Ohishi, ``Motion copying system based on
  real-world haptics in variable speed,'' in {\em Proceedings of 2008 13th
  International Power Electronics and Motion Control Conference},
  pp.~1604--1609, 2008.

\bibitem{chinthaka18:_inert_compen_motion_copyin_system}
M.~D. Chinthaka and T.~Shimono, ``Inertia compensation of motion copying system
  for dexterous object handling,'' {\em IEEJ Journal of Industry Applications},
  vol.~7, no.~6, pp.~495--505, 2018.

\bibitem{Tsuji2016}
T.~Tsuji, J.~Ohkuma, and S.~Sakaino, ``{Dynamic Object Manipulation Considering
  Contact Condition of Robot with Tool},'' {\em IEEE Transactions on Industrial
  Electronics}, vol.~63, no.~3, pp.~1972--1980, 2016.

\bibitem{tsuji18:_optim_trajec_gener_model_predic}
T.~Tsuji, K.~Kutsuzawa, and S.~Sakaino, ``Optimized trajectory generation based
  on model predictive control for turning over pancakes,'' {\em IEEJ Journal of
  Industry Applications}, vol.~7, no.~1, pp.~22--28, 2018.

\end{thebibliography}

\end{document}